# Towards edible drones for rescue missions: design and flight of nutritional wings

Bokeon Kwak[1*], Jun Shintake[2*](*equal contribution*), Lu Zhang[3], and Dario Floreano[1]

[1] Bokeon Kwak, and Dario Floreano are with the Laboratory of Intelligent Systems, School of Engineering, Ecole Polytechnique Federale de Lausanne, CH1015 Lausanne, Switzerland (e-mail: bokeon.kwak@epfl.ch; dario.floreano@epfl.ch).
[2] Jun Shintake is with the Shintake Research Group, School of Informatics and Engineering, The University of Electro-Communications, 1-5-1 Chofugaoka, Chofu, Tokyo 182-8585, Japan (e-mail: shintake@uec.ac.jp).
[3] Lu Zhang is with Laboratory of Food Process Engineering, Wageningen University and Research, P.O. Box 17, 6700 AA Wageningen, the Netherlands (e-mail: lu1.zhang@wur.nl).
*Bokeon Kwak and Jun Shintake contributed equally to this work.

*Abstract*—Drones have shown to be useful aerial vehicles for unmanned transport missions such as food and medical supply delivery. This can be leveraged to deliver life-saving nutrition and medicine for people in emergency situations. However, commercial drones can generally only carry 10 % − 30 % of their own mass as payload, which limits the amount of food delivery in a single flight. One novel solution to noticeably increase the food-carrying ratio of a drone, is recreating some structures of a drone, such as the wings, with edible materials. We thus propose a drone, which is no longer only a food transporting aircraft, but itself is partially edible, increasing its food-carrying mass ratio to 50 %, owing to its edible wings. Furthermore, should the edible drone be left behind in the environment after performing its task in an emergency situation, it will be more biodegradable than its non-edible counterpart, leaving less waste in the environment. Here we describe the choice of materials and scalable design of edible wings, and validate the method in a flight-capable prototype that can provide 300 kcal and carry a payload of 80 g of water.

## I. INTRODUCTION

Unmanned aerial vehicles (UAVs), commonly referred to as drones, are versatile aircrafts that can perform various tasks, including the delivery of small parcels, food, and medical supplies. Some companies have already launched drone delivery services to reduce the cost of delivering small items on the last mile [1]. In this case, multirotor-type drones are most commonly adopted, owing to their reliability when hovering and maneuvering. Drones can also be used to deliver life-saving nutrition for people in emergency situations, where approaches for ground vehicles are difficult [2]. Here, high UAV endurance and large payload are necessary to reach a remote place with enough food to sustain the endangered person. Consequently, for this scenario utilizing a fixed-wing drone is advantageous over a multirotor-type drone [3]. Nevertheless, most fixed-wing drones can carry only 10 % − 30 % of their own mass as payload [3], [4]. To noticeably increase the ratio of food-carrying mass to total drone mass, a new method of food mobilization is necessary.

In this study, we present a solution for the problem, which recreates body structures of a drone with food materials without sacrificing mechanical properties. As illustrated in Fig. 1a, we envisioned that a (partially) edible drone could be deployed to a remote place and be ingested by a person who needs emergency food. Hence an edible drone would be consumable and designed to perform only a one-way trip. Given that high-speed edible motors and associated electronics are not yet available, we focused on replacing the structural section of the fixed-wing drone, i.e. the wing. In general, the volume of the wing occupies the largest part of a fixed-wing drone. We therefore hypothesized that making this major component of a UAV from edible materials would itself allow us to significantly increase the drone's inherent food-carrying mass ratio compared to a non-edible drone (refer Fig. 1b). Owing to this high ratio, an edible drone is a more cost-effective way to deliver target nutrients than existing drones. As a proof of concept, the edible-winged drone proposed in this study has a wingspan of 678 mm and can therefore provide 300 kcal of nutrition. In addition, shifting the nutrients to the wing structure frees the payload for carrying liquids, and the proposed drone can theoretically carry 80 g of water. Considering a one-way trip and the possibility that the drone may not be entirely eaten, it will leave less waste in the environment because it will be more biodegradable than its non-edible counterpart.

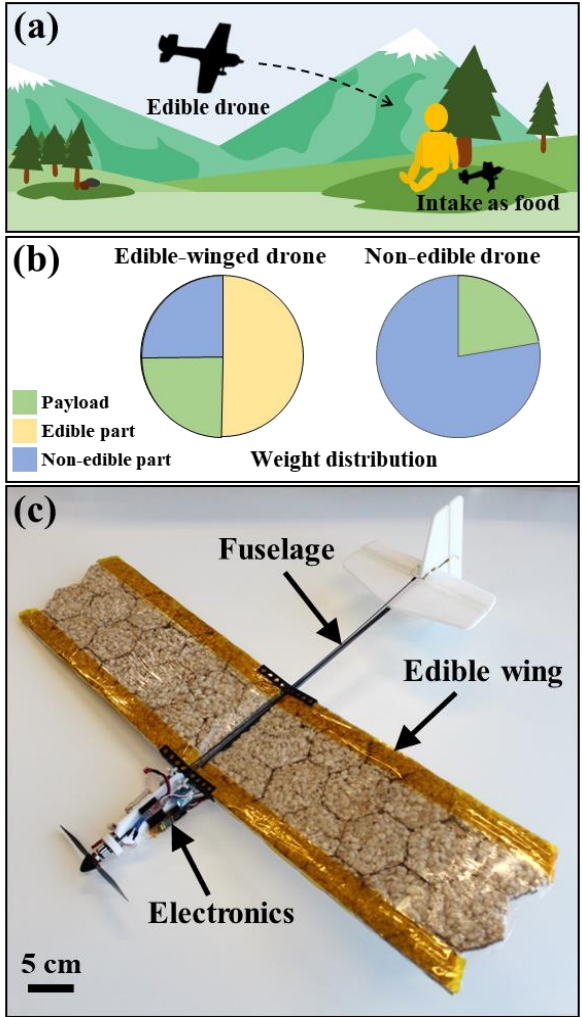

Figure 1. The concept of an edible-winged drone, designed to provide lifesaving nutrition in a remote place. (a): deployment of edible drone, (b): higher food-carrying ratio of edible-winged drone over non-edible drone, (c): the first partially edible drone, featuring rice cookie wings (scale bar: 5 cm).

The idea of using edible or biodegradable material in robotics has already been considered in a number of previous works. Development of soft actuators using edible or biodegradable materials is particularly well studied topic [5]–[12]. Among them, gelatin was a widely chosen material, owing to its softness, low-cost, and ease of processing [6]–[8], [10]–[12]. Other materials such as living natto cells [5] and popcorn kernels [9] were also considered as actuation sources. Besides actuators, edible/biodegradable materials have previously been utilized to make robotic bodies; a self-propelling boat using edible wax and gelatin [13], a fermented robot made of vegetables [14], and an ingestible wound treating robot [15] are examples. Another growing field gaining attention is edible electronics and its application to robotics [16]. One recent study proposed a cellulose=based strain sensor, which could potentially be used in a flying wing [17]. Despite many previous developments in edible robotics, however, an edible aerial system has not yet been studied. Additionally, previous studies did not aim at providing humans with nutritional energy. As a first effort to fill the gap, this work provides a scalable design process of a fixed-wing edible drone (refer Fig. 1c) and its outdoor flight demonstration.

The rest of this paper is organised as follows. The material selection suitable for an edible wing is discussed in section II, followed by the scalable design process of the proposed edible drone in section III. The fabrication process of the edible drone and its flight test are given in section IV. Lastly, the significance and limitation of the edible drone, and required future development are discussed in section V.

## II. MATERIALS FOR EDIBLE WINGS

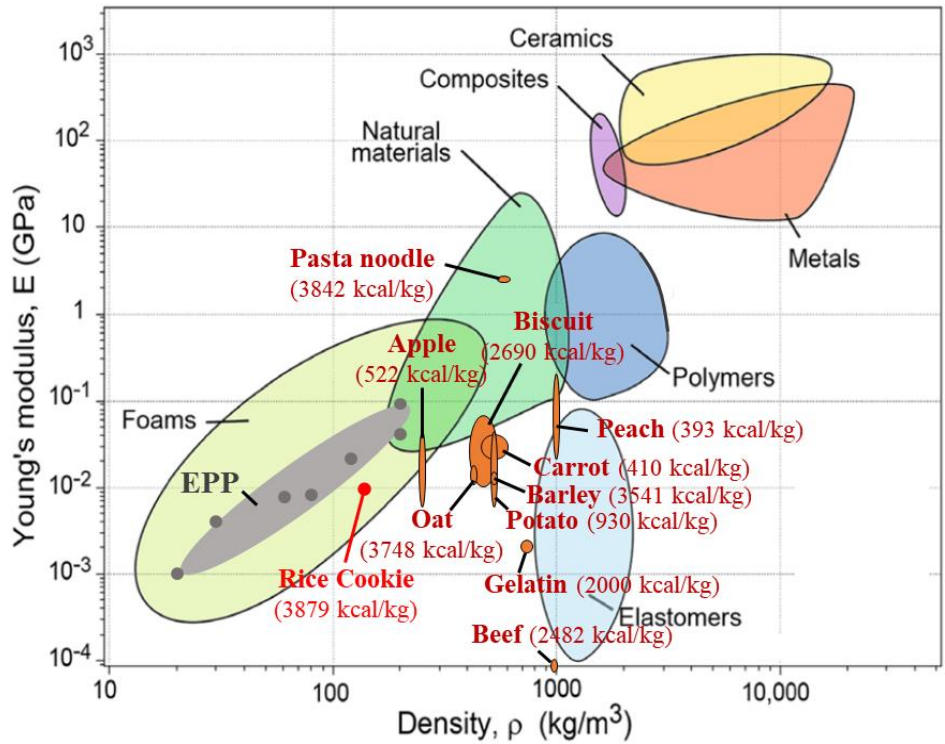

Figure 2. Fig. 2. Comparison of Young's moduli ($E$) and densities ($\rho$) of engineering materials and food materials. The properties of engineering materials were reproduced from the Ashby chart [18] and other literature [19]–[21]. The properties of a rice cookie were measured by the authors. The rest of the food material properties were obtained from [22]–[27]; the orange bubbles show the range of $E$ and $\rho$ for a given material. Generally, $E$ and $\rho$ of food materials are located between that of foam and elastomer. The calories per unit mass of the food materials are also given based on the calorie data in [28].

As wings occupy the largest volume of a fixed-wing drone, replacing the wings with food materials can provide a significantly higher nutrition-carrying mass ratio compared to a non-edible drone, which is pertinent to rescue missions. In this study, the wings were designed to be edible, whereas the remaining structures (e.g., fuselage, electronics, etc.) were built using conventional materials. An edible wing proposed in this study was fabricated by gluing main structural materials with edible adhesive. Thus, the main material is addressed first, followed by the selection of an edible adhesive, which can provide the strongest adhesion to the selected main material.

### A. Main structural material of edible wing

Just like many existing wings, an edible wing should have a low density and high Young's modulus for stable and efficient flight. Lowering the density of an edible wing may reduce the food-mass ratio of an edible drone; however, it is compensated by the large volume of the edible wing itself. Maintaining a low mass is the foremost requirement for all scales of aircraft design. At the same time, the wing should be strong enough to avoid bending or material failure during flight. These requirements favor using a foam, such as expanded polypropylene (EPP) as a primary structural material for conventional fixed-wing drones. The Young's modulus E and density $\rho$ of EPP exhibit high variance depending on the EPP foam manufacturers, but they are positively correlated as shown in Fig. 2. These mechanical properties were collected from the literature [19]–[21] and depicted on Ashby's material property plot [18], which concisely shows $E$ and $\rho$ of common engineering materials.

To identify a food material, which has the closest mechanical properties to EPP, $E$ and $\rho$ of different food materials were also depicted on Fig. 2. The FAO-INFOODS database was exclusively used to get densities [22], while Young's moduli of the relevant food materials were obtained from other literature [23]–[27]. As the number of food materials with known $E$ and $\rho$ are limited, some representative solid state foods were selected to give an indication of their mechanical properties. Overall, the $E$ and $\rho$ of food materials were positioned between the foams and elastomers. Nevertheless, for a suitable wing material, we needed to find a food whose $E$ and $\rho$ were close to the EPP. One of the most promising candidates was a puffed rice cookie (off-the-shelf, Naturaplan), of which the $E$ and $\rho$ were measured using a 3-point-bending test and mass of sample of known size, respectively. The measurement results were $E = 10.4 \pm 1.3$ MPa and $\rho = 112 \pm 8.4$ kg/m3. A rice cookie is produced by applying high pressure to rice grains at high temperature, which puffs the rice grains and lowers the $\rho$ significantly compared to other food materials. Besides the similarity of $E$ and $\rho$ to EPP, a rice cookie is also easily machinable by laser cutting. Consequently, we chose puffed rice cookies as a primary structural material for the edible wing in this work.

Maximizing the number of calories per unit mass is also essential to ensuring that an edible drone can efficiently perform future rescue missions. In the case of a rice cookie, it has 3870 kcal per 1 kg (refer to Fig. 2), according to the manufacturer's data. This is lower than the number of calories of some sweets (e.g., over 5000 kcal/kg for chocolate and candy [28]). However, the densities of those sweets are $5 \times 8$ times higher than that of the rice cookie [22], and therefore they are not suitable for prototyping an edible wing. Rice cookies offer a very similar nutritional value (kcal/kg) to other common foods, such as oats, barley, and pasta (Fig. 2), however rice cookies are less dense and hence more suitable as a material for the edible-winged drone.

*B. Selection of edible adhesive*

(a)
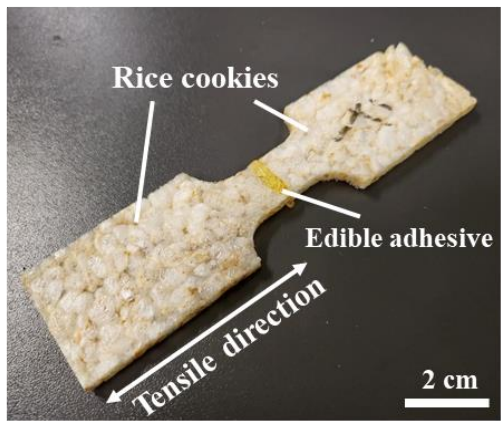

(b)
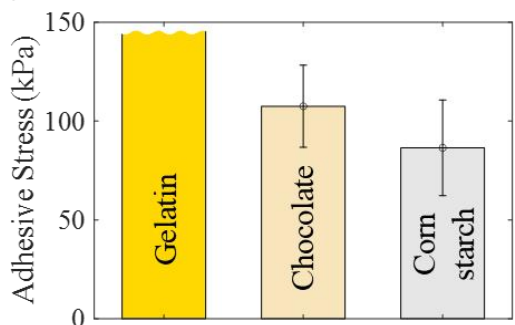

Figure 3. Comparison of the adhesive stress of three edible adhesives using a tensile pulling test (a: specimen; scale bar: 2 cm, b: measurement result).

The typical size of a rice cookie purchased from a supermarket is around 70 mm (in length). An edible wing has a much larger area, which means that multiple rice cookies need to be laser cut (to maximise adhesive surface) and connected by using an edible adhesive. In this study, we tested three types of edible adhesives; namely, corn starch, chocolate, and gelatin. Corn starch (Patissier) glue was prepared by mixing corn starch powder and warm water (60°C) in a 1:1 ratio. Chocolate pellets (Callebaut) were simply microwaved (1000 W, 20 s) before use as an adhesive. In the case of gelatin glue, gelatin powder (Sigma Aldrich) and warm water (60°C) was mixed in a 1:3 ratio. These edible adhesives also provide a small amount of nutrients to the edible wing.

To quantitatively measure their adhesive strengths, at least seven samples were prepared for each type of edible adhesive. The size and shape of the specimens were determined by referring to the adhesion measurement sample shown in [29]. Note from Fig. 3a that two laser cut rice cookies were bonded by one of the three edible adhesives. It was important to dry the bonded sites at least 12 hours to allow any water residue to evaporate. Each sample was then fixed to a universal material tester (Instron 5965), and tensile stress (rate: 0.08 mm/s) was applied until breakage of the bonded site. The maximum tensile force divided by the cross section of the bonded site (i.e., adhesive stress) was finally obtained and depicted in Fig. 3b for comparison. In the case of gelatin, for 10 consecutive test, the bonded site was never broken, but the rice cookie itself failed after 1.6% extension. In other words, gelatin maintained a strong bond until material failure of the rice cookie itself. The maximum tensile stress measured before the material failure was 150 kPa; the gelatin adhesive can withstand even higher tensile stress than 150 kPa as long as the rice cookie is not broken. Although it was difficult to quantify the adhesive stress of gelatin, we could conclude that it was stronger than both corn starch and chocolate; the former was 79.4±18.3 kPa and the latter was 113.3±15.1 kPa. Therefore, gelatin was used as an edible adhesive throughout the study.

## III. SCALABLE DESIGN OF THE EDIBLE-WINGED DRONE

### A. Design requirements

This section describes the scalable design process of the edible wing where the payload is reserved for water. In this study, the mass of payload, drone without the payload, and drone with the payload was denoted as $W_{\text{payload}}$, $W_{\text{wo payload}}$, and $W$ (i.e., $W = W_{\text{payload}} + W_{\text{wo payload}}$), respectively. Here, we introduce one example of an edible drone carrying 300 kcal as an edible wing and 80 g of payload (e.g. water) as $W_{\text{payload}}$. Note that 300 kcal is equivalent to the calorie intake from an average breakfast [30]. By following the scalable design process of this work, a designer could load other amounts of nutrition onto the drone, depending on their design requirements. The $W_{\text{payload}}$ of fixed-wing drones scales with their size [4], which means that delivering a higher payload, such as nutrition or water, will increase the size of the edible drone. According to statistical data collected from [4], [31]–[35], the ratio $W_{\text{payload}}/W$ was found to be 0.272 as shown in Fig. 4a. By setting $W_{\text{payload}}$ to 80 g, the resultant $W_{\text{wo payload}}$ became 214 g, and therefore $W$ = 294 g. This mass estimation was used in the rest of the design process.

### B. Size of edible wing

After an initial mass estimation, the edible wing itself needed to be designed. For simplicity, wing sweep or taper were not considered. Instead, the wing was modeled as a flat plate (thickness: $t$) as illustrated in Fig. 4b, as this edible wing was to be fabricated with rice cookies. The aerodynamic performance of a flat plate is generally invariant to its chord Reynolds number ($Re$), with better performance achieved as $t$ is decreased [36]. Thus, we limited $t$ to the thickness of a single layer rice cookie. Note that the chord length ($c$) was used to evaluate $Re = \rho V c/\mu$ throughout this paper where $\rho$, $V$, and $\mu$ were air density (1.225 kg/m$^3$), characteristic velocity, and air viscosity (1.81 × 10$^{-5}$ kg/m · s), respectively.

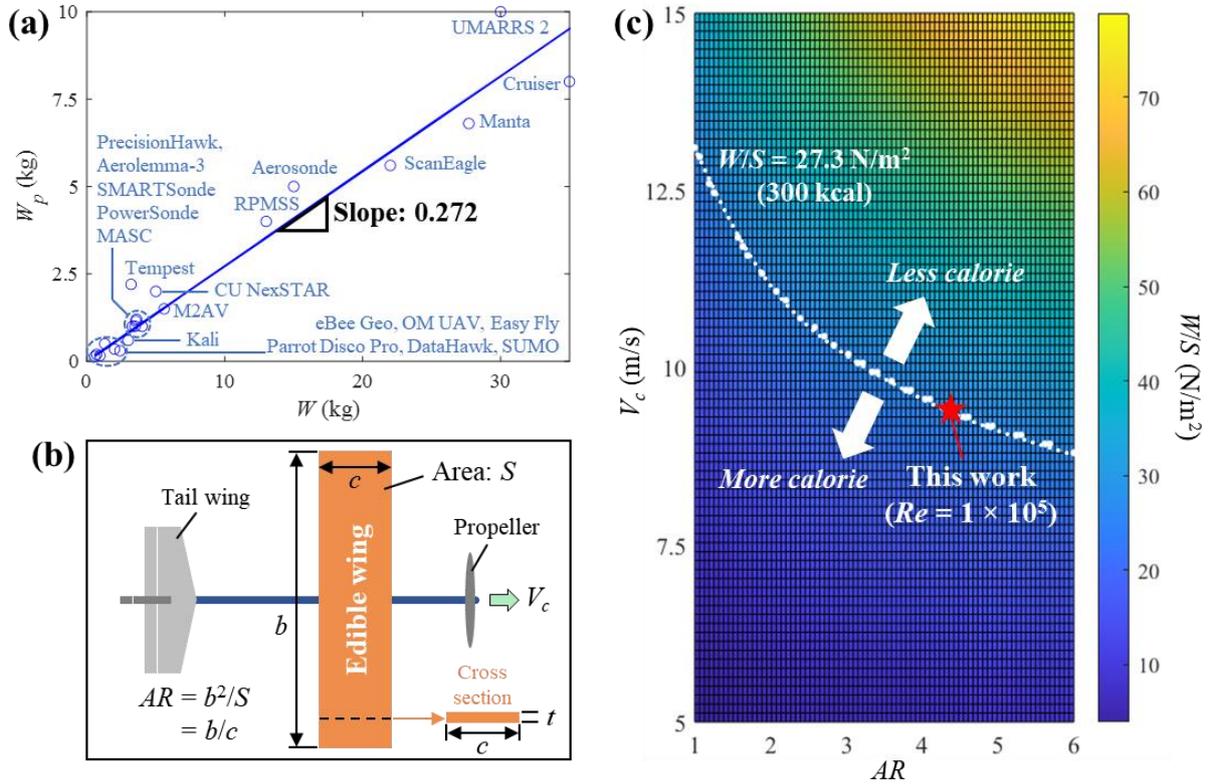

Figure 4. Design of an edible-winged drone for a given number of calories and payload. (a): statistical data from previous fixed-wing drones ($W_{payload}$: payload, $W$: gross mass), (b): simplified illustration of an edible-winged drone in top view ($b$: wingspan, $c$: chord, $t$: thickness, $AR$: aspect ratio, $V_c$: cruise speed), (c): the distribution of wing loading ($W/S$) with respect to $V_c$ and $AR$ where the white dotted line indicates the design space of the edible wing to allow it to have 300 kcal.

As the nutritional content to be carried had already been decided, the reference area of the edible wing ($S$) could be calculated. According to the nutritional data provided by manufacturers, 100 g of rice cookie provides 387 kcal. In the case of the gelatin adhesive, its calorie content was assumed to be 200 kcal per 100 g, according to the calorie data of a commercially available gelatin-based adhesive [37]. In our preliminary test, we found that laser-cut rice cookies and edible adhesive were used in 4:1 mass ratio to concatenate into a flat plate of given area. By considering the kcal per gram data of each food material, we knew that 28.4 kcal would be contained in 100 cm² of edible wing. As a result, $S$ was calculated to be 1056.34 cm² in order for the wing to contain 300 kcal. Since we estimated $W$ as 294 g (2.88 N) in the previous section, the associated wing loading ($W/S$) was found to be 27.3 N/m². Note that wing loading is a critical design parameter in aircraft design, which highly impacts the lift coefficient, stall speed, and nutrition that it is able to carry in the case of an edible drone. During the cruise flight of a propeller-powered aircraft, the following constraint can be considered:

$$\frac{W}{S} = \frac{1}{2}\rho V_c^2 \sqrt{\pi e AR C_{D0}} \tag{1}$$

$$e = 1.78(1 - 0.045 AR^{0.68}) - 0.64 \tag{2}$$

Here $e$ is the Oswald span efficiency factor, $AR$ is the aspect ratio of the wing, and $C_{D0}$ is the zero-lift drag coefficient and is assumed to be 0.02 for a propeller aircraft [38]. The distribution of $W/S$ with respect to $V_c$ and $AR$ is given in Fig. 4c, where the goal $W/S$ = 27.3 N/m² is highlighted with a white dotted line and is associate with 300 kcal nutritional content as discussed earlier; each point of the white dotted line represents a unique size of the edible wing. Note that Fig. 4c is scalable with respect to calorie content. More precisely, when the amount of nutrition to be delivered is changed, *S* will be adjusted accordingly. For example, when the designer wants to increase the number of calories, the resultant white dotted line will move downwards and vice versa.

*Table 1. DESIGN PARAMETERS OF THE EDIBLE WING. NUTRITION TO CARRY, $W_{payload}$, Re, AND α ARE VALUES DETERMINED AT THE START OF THE DESIGN PROCESS.*

| Design parameters | Values |
|---|---|
| Nutrition to carry | 300 kcal |
| Payload ($W_{payload}$) | 80 g |
| Estimated emptied drone mass ($W_{wo\ payload}$) | 214 g |
| Gross mass of drone ($W$) | 294 g |
| Wing reference area ($S$) | 1056.3 cm$^2$ |
| Wing loading ($W/S$) | 27.3 N/m$^2$ |
| Wing aspect ratio ($AR$) | 4.3 |
| Wing chord ($c$) | 155.9 mm |
| Wingspan ($b$) | 678.8 mm |
| Cruise speed ($V_c$) | 9.43 m/s |
| Chord Reynolds number ($Re$) | $0.996 \times 10^5$ (~ $1 \times 10^5$) |
| Angle of attack ($\alpha$) | 7.2° |
| Lift coefficient ($C_L$) | 0.5036 |
| Stall speed ($V_s$) | 9.41 m/s |

For a given $W/S$, the designer must choose one particular $AR$ to prototype an edible wing. In this study, we focused on one particular Reynolds number $Re = 1 \times 10^5$, for the following reasons: a thin plate is more efficient than a conventional airfoil below a Reynolds number of $Re = 1 \times 10^5$, while a conventional airfoil performs best at $Re > 10^6$ [39]. This happens because of the combined effect of early flow separation from an airfoil and late reattachment, which occurs near the trailing edge when $0.5 \times 10^5 < Re < 1 \times 10^5$ [36]. Therefore, $Re = 1 \times 10^5$ can be considered as the upper limit for a thin plate to function as a wing. By establishing such a $Re$ constraint, the exact size of the edible wing was finally obtained (namely, $AR$ = 4.354 where $b$ = 678.8 mm and $c$ = 155.9 mm). Since $V_c = Re \cdot \mu/(\rho c)$, the resultant cruise speed was $V_c$ = 9.43 m/s. Furthermore, the designed $W/S$, $V_c$, and $W$ are all consistent with the Great Flight Diagram by Tennekes [40], which concisely shows the characteristics of animal flight and of aircrafts of different scales. The summary of design parameters for the edible-winged drone presented in this paper are tabulated in Table 1.

We also calculated the critical angle of attack ($\alpha$), which ensures that $V_c$ = 9.43 m/s is higher than the stall speed ($V_s$). Note that $V_s$ is the minimum flight speed and needs to be maintained and obtained from:

$$\frac{W}{S} = \frac{1}{2}\rho V_s^2 C_{L\ max} \qquad (3)$$

where $C_{L\ max}$ is the maximum lift coefficient of an edible wing [38]. Generally, more lift can be generated at higher angle of attack. The lift ($C_L$) generated by a rectangular wing was calculated from the Lowry & Polharmus model given below [41]:

$$C_L = \frac{2\alpha\pi AR}{2 + \sqrt{4 + AR^2}} \qquad (4)$$

Note that $W/S$ = 27.3 N/m$^2$ and $AR$ = 4.35 in (3) and (4), respectively. By assuming $C_{L\ max}$ is equivalent to the $C_L$ obtained from (4), the smallest critical $\alpha$, which satisfies $V_c > V_s$, is 7.2°; in this case, $V_s$ was 9.41 m/s. Plugging in $AR$ = 4.35 and $\alpha$ = 7.2° into (4) yielded $C_L = C_{L\ max}$ = 0.5036. This result was similar to $C_L$ = 0.413 obtained from the Anderson model [42]. Additionally, the experimentally measured $C_L$ of a flat plate at $α = 7.2°$ was 0.549 when $Re = 1 × 10^5$ [41]. Thus, letting $C_{L\,max} = 0.5036$ obtained from (4) was a reliable estimation when $α = 7.2°$. The highest lift-to-drag ratio ($L/D$) of a thin flat plate operating at $Re = 1 × 10^5$ was 7 when $α = 5°$ [41]. At the same $Re$ but with $α = 7.2°$, $L/D = 6.2$, which means that the edible-winged drone can theoretically achieve 88.6% of the maximal $L/D$ generated by a thin plate wing model.

*C. Thrust to weight ratio*

Thrust-to-weight ratio ($T/W$) is another dimensionless parameter in aircraft design which quantifies the ratio of thrust ($T$) to $W$. Higher $T/W$ accelerates a drone more rapidly; however, an overly large $T/W$ requires a larger motor and battery power. Therefore, $T/W$ of an edible drone should be appropriately adjusted. The required $T/W$ for cruise flight can be calculated as follows:

$$\frac{T}{W} = \frac{qC_D S}{W} + \frac{1}{\pi ARqe}\frac{W}{S} \qquad (5)$$

where $C_D$ is the drag coefficient of the edible-winged drone, and $q = 0.5\rho V_c^2$ is the dynamic pressure during cruise flight [43]. By considering previously reported $C_D$ values and the experimentally measured drag coefficient of a flat plate [42]–[44], $C_D$ was chosen to be 0.045, and the resultant $T/W$ from (5) became 0.130. This value was similar to the $T/W$ obtained from the thrust matching condition for cruise flight (i.e., $T/W = 1/(L/D)$) [38]; the inverse of lift-to-drag ratio ($L/D$) of a flat plate at $Re = 1×10^5$ and $α = 7.2°$ was 0.161 [41]. The maximum thrust generated by the motor was 1.079 N, which yielded the maximum $T/W$ of the edible-winged drone as 0.374. As the maximum $T/W$ of the edible-winged drone was ~ 2 to 3 times larger than the theoretical $T/W$ values, it was likely that the designed drone would be able to fly by not necessarily actuate the motor at full throttle.

*D. Tail wing design*

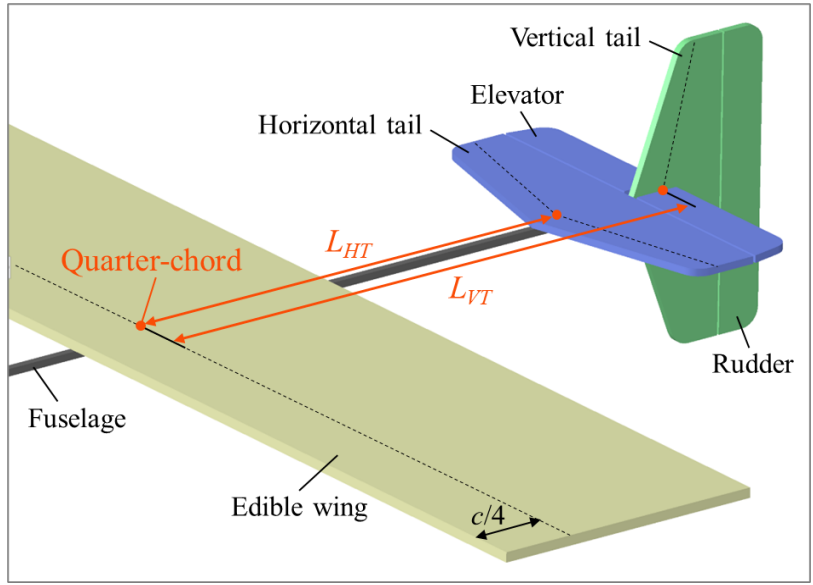

Figure 5. The illustration of the tail wing. The red dots denote the quarter-chord of the edible wing and the tail wing, which were used to determine the size and position of the tail wing.

In this study, the tail wing was fabricated using an inedible foam to focus our study to edible wings in isolation without the influence of other conventional parts. Two types of control surfaces (elevator and rudder) were employed for maneuvering as shown in Fig. 5. When determining the size and position of the tail wing, a vertical tail volume coefficient ($c_{VT}$) and horizontal tail volume coefficient ($c_{HT}$) were used, which are defined as shown below:

$$C_{VT} = \frac{L_{VT}S_{VT}}{bS}, \quad C_{HT} = \frac{L_{HT}S_{HT}}{cS} \quad (6)$$

where $L_{VT}$ and $L_{HT}$ were the distance between the edible wing quarter-chord and the tail quarter-chord (refer Fig. 5) [38]. $S_{VT}$ and $S_{HT}$ were the vertical tail area and horizontal tail area, respectively. Generally, $c_{HT}$ was designed to be 5 to 12 times larger than $c_{VT}$ [38], [44]. We proceeded by predetermining the size and shape of the tail wing first, then continued adjusting the $L_{VT}$ and $L_{HT}$ to obtain $c_{VT} = 0.05$ and $c_{HT} = 0.25$. These coefficients were determined by considering the fuselage length and were subsequently shown to be suitable values from the flight test.

## IV. FABRICATION AND FLIGHT TEST

As already addressed in section II, the materials for the edible wing were chosen to be puffed rice cookies, connected using gelatin adhesive. Hexagonal tiling has been proven to divide a surface into regions of equal areas with the lowest total perimeter [45]. Thus, a hexagonal pattern was chosen to minimize the additional mass added by the edible adhesive. The rice cookies were laser cut into either hexagonal or half-hexagonal (isosceles trapezoidal) shapes and glued with gelatin to create a planar structure, as illustrated in Fig. 6a. Subsequently, the entire wing was dried at room temperature for 12 hours. The wing was then divided into two halves and each half further processed separately. One piece of the edible wing (i.e., half-wingspan sized) is shown in Fig. 6b. Here, the measured $c$ and $b/2$ were 150 mm and 340 mm, respectively displaying less than 3% fabrication error compared to the designed values in Table I. To prevent any humidity damage, the entire wing surface was wrapped in plastic film and tape. The mass of one complete edible wing (full wingspan), including the protection film, was 100 g.

During the level flight of the edible-winged drone with a full payload, a lift generated from the edible wing must be equal to its own weight $W_g$, where $g = 9.81$ m/s$^2$ is the gravitational acceleration constant. If the edible wing cannot withstand such $Wg$ amount of lift, then the wing will be broken. Thus, a strength test was also performed to ensure each half-wingspan edible wing can withstand $0.5Wg$ of lift. To simulate the $0.5Wg$ of lift applying to a half-wingspan ($b/2$) edible wing, a test bed was prepared as shown in Fig. 6c. Here, a half-wingspan edible wing was connected to the fixture (blue structure), and the resultant deflection at the tip ($\delta$) was measured upon force distribution $F(x)$. By letting $L_s = \int_0^{b/2} F(x)\, dx$ as a simulated lift force to the half-wingspan edible wing, its structural integrity is guaranteed when the half-wing can at least withstand $L_s = 0.5Wg$ or higher. In this study, Schrenk's approximation was used to simulate the spanwise force distribution [38]:

$$F(x) = \frac{1}{2}\left(\frac{wg}{b} + f_0\sqrt{1 - \left(\frac{x}{0.5b}\right)^2}\right) \quad (7)$$

where $f_0$ was the y-axis intercept of $F(x)$. Three different $L_s$ were applied to the half-span wing by stacking granule-filled bags onto the wing surface, and the resultant $\delta$ was measured. As shown in Fig. 6c, the wing exhibited 1.2 mm of deflection on average when no load was applied. When $L_s$ was 1.04 N and 1.56 N, the associated $\delta$ was 3.7 mm and 5.9 mm on average, respectively. However, the half-span wing was broken upon applying $L_s = 2.09$ N; meaning that the maximum $L_s$, which can be supported by the half-wing was between 1.56 N and 2.09 N. Nevertheless, the structural integrity of the edible wing was verified as it could withstand $L_s = 0.54Wg$. This implied that a full-span edible wing could withstand $1.08Wg$ of lift. As a result, the proposed edible wing was verified to withstand the lift required to maintain the level flight of the edible-winged drone with full payload. Furthermore, by considering $0.54Wg = 0.75W_{\text{wo payload}}g$ the proposed wing could also withstand 1.5 times the lift required to maintain the level fight of an edible-winged drone without payload.

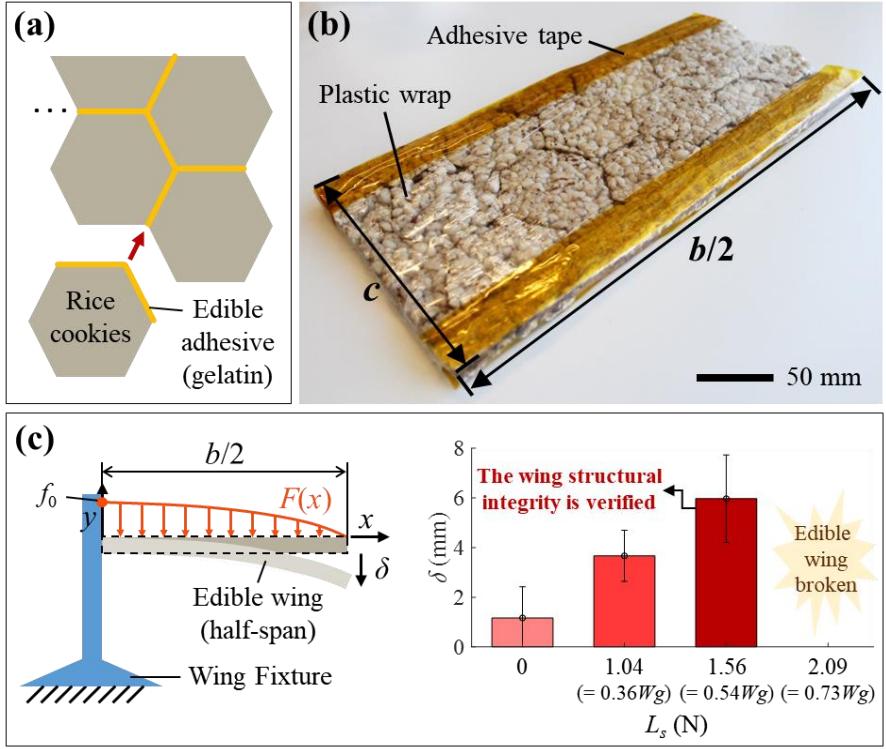

Figure 6. Fabrication method of the edible wing and its strength test. (a): gluing of hexagonal shaped rice cookies, (b): the edible wing was wrapped in plastic film for humidity protection, where $c = 150$ mm and $b/2 = 340$ mm from the prototype; scale bar: 50 mm, (c): the deflection ($\delta$) of the edible wing under distributed load $F(x)$, where a simulated lift force $L_s = \int_0^{b/2} F(x)\,dx$. The edible wing is fixed to a wing fixture (blue structure) during experiment.

The proposed edible-winged drone is shown in Fig. 1c. The fuselage was made of a 0.5 m long hollowed carbon rod, which was chosen based on excellent material strength. The edible wing was mounted to the fuselage with a 10° dihedral angle for stable flight. All of the electronics were located at the frontal side for stable mass balance during the flight. Here, the brushless motor (AP05 3000kv), equipped with a 5030 propeller, was used to generate thrust. In addition, two micro servo motors (Blue Arrow, torque: 0.8 kg-cm) were employed to actuate the elevator and rudder of the tail wing (recall Fig. 5). All of the motors were remotely controlled by an ordinary 2.4 GHz RC (radio control) transmitter and receiver set. The edible-winged drone was powered by an 18.2 g Li-Po battery (7.4 V, 260 mAh) for at least 10 minutes of flight. The entire mass without payload was 200 g, which was very similar to our initial estimation $W_{wo\ payload} = 214$ g, given in section III.A. Furthermore, the mass percentage of the edible part of the drone without payload was 50 %. Note that the current edible-winged drone did not have a container to carry 80 g of payload ($W_{payload}$), such as water; instead, this condition was only theoretically considered, allowing us to focus on the first flight tests of the edible-winged drone. In outdoor tests, the drone displayed stable flight at a cruise speed of 9.87 m/s (Fig. 7, see also supplementary video), which is similar to the theoretical cruising speed of 9.83 m/s, computed in section II.B. Future development will focus on a novel way to store payloads, such as water, on an edible drone, without significantly increasing the surface area (exposed to air).

## V. DISCUSSION AND CONCLUSION

Development of edible systems has recently gained popularity in robotics research because of the material sustainability and novel applications [46]. The edible-winged drone described in this paper is the first aerial system (partially) made of edible material, proposing a novel food mobilization method for delivering life-saving nutrition. Until now, the amount of food that existing drones could carry was restricted to the payload. However, an edible drone can distinctly overcome this payload limitation, owing to the recreation of some body structures with food materials. A scalable design method is studied, which enables the designer to embed a specific amount of nutrition and water into a drone. The ability of the proposed edible-winged drone to fly is also shown, which demonstrates its future applicability to rescue missions, where emergency nutrition can be delivered to people in need. Owing to the facile fabrication method of this prototype, multiple edible drones could be easily fabricated and deployed, allowing for the delivery of much higher amounts of nutrition.

This study shows that it is feasible to fabricate a drone, the body of which contains nutrients that could be delivered to areas with difficult access in emergency situations. This approach enables an edible-winged drone to carry more water as a payload. Additionally, the scalability of the proposed design can load other nutritional values required by an endangered person. However, further research effort is needed to i) further increase the total calorie content that can be provided (by developing an edible fuselage, edible tail wing, etc.) and improve the nutrient composition (i.e., ratios of macro-nutrients such as protein, carbohydrates and lipids) to meet the requirements defined by WHO for human nutrition; ii) store additional payload, such as water, inside an edible container without significantly increasing the surface area exposed to the air; iii) investigate and incorporate edible sensors or edible electronics to increase the mass ratio of the biodegradable/edible part of the drone; iv) explore other methods to fabricate the wing or drone, for example, by 3D printing, to improve fabrication efficiency and reduce time.

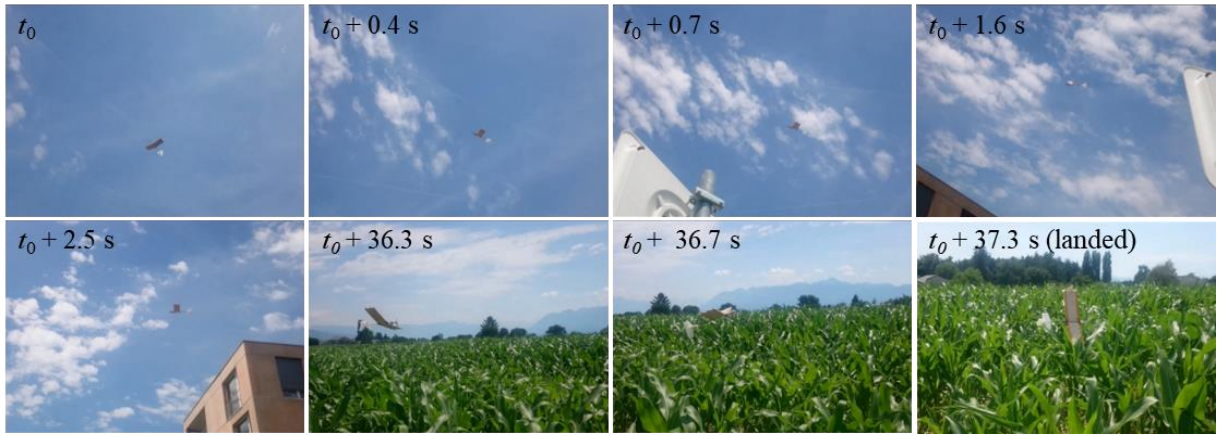

Figure 7. Selected footage from the flight test of the edible drone ($t_{0I}$: reference time). Supplementary video is available online.

## VI. ACKNOWLEDGMENT

The authors are grateful for the technical help and discussion provided by R. Arandes, S. Zhang and W. Stewart. They also wish to thank M. Pankhurst for her help in editing the manuscript. This work was supported by the European Union's Horizon 2020 research and innovation program under Grant agreement 964596 ROBOFOOD.